\newif\ifcomments  
\commentsfalse

\documentclass[a4paper,table]{article}
\usepackage{INTERSPEECH2021}
\usepackage[dvipsnames]{xcolor}
\usepackage{highlight}
\usepackage{comment}
\usepackage{hyperref}
\usepackage{url}
\usepackage{graphicx}
\usepackage{enumitem}

\title{Lookup-Table Recurrent Language Models for Long Tail Speech Recognition}
\name{W. Ronny Huang, Tara N. Sainath, Cal Peyser, Shankar Kumar, David Rybach, Trevor Strohman}

\address{Google Research}
\email{\{wrh, tsainath, cpeyser, shankarkumar, rybach, strohman\}@google.com}

\begin{document}
\maketitle
\begin{abstract}
We introduce Lookup-Table Language Models (LookupLM), a method for scaling up the size of RNN language models with only a constant increase in the floating point operations, by increasing the expressivity of the embedding table.
In particular, we instantiate an (additional) embedding table which embeds the previous n-gram token sequence, rather than a single token. This allows the embedding table to be scaled up arbitrarily---with a commensurate increase in performance---without changing the token vocabulary.
Since embeddings are sparsely retrieved from the table via a lookup; increasing the size of the table adds neither extra operations to each forward pass nor extra parameters that need to be stored on limited GPU/TPU memory.
We explore scaling n-gram embedding tables up to nearly a billion parameters. 
When trained on a 3-billion sentence corpus, we find that LookupLM improves long tail log perplexity by 2.44 and long tail WER by 23.4\% on a downstream speech recognition task over a standard RNN language model baseline, an improvement comparable to a scaling up the baseline by 6.2x the number of floating point operations.
\end{abstract}
\noindent\textbf{Index Terms}: language modeling, speech recognition, embedding, n-gram, long tail

\section{Introduction}
Improving the decoding of rare words or sequences is an ongoing problem in natural language tasks. 
For example, speech recognition systems misrecognize many legitimate token sequences that appear with zero or low frequency in transcribed acoustic data. This is particularly a problem with end-to-end (E2E) models \cite{graves2012sequence, chan2015listen}, which are typically trained on a small-fraction of audio-text pairs compared to the amount of text-only data. In addition, E2E models tend to have narrower beam searches which also contributes to poor rare-word performance \cite{zhao2019shallow}.
One solution to this problem is to integrate language models trained on text-only data which often contain rich long tail content \cite{peyser2020improving}.
For example, a language model trained on a corpus containing rare words absent in speech data can bias a speech recognition system toward correctly decoding those rare words via language model fusion \cite{kannan2018analysis, toshniwal2018comparison, sriram2017cold, weng2020minimum}. Note we use ``rare word'' and ``long tail'' interchangeably throughout this paper to mean rare sub-sequences which may or may not occupy an entire example.

Neural language models have shown in recent years that size matters. Given sufficient compute and data, our ability to model the vast complexity of language scales with the size of the model \cite{kaplan2020scaling}, with perplexities showing no sign of saturating even up to a trillion parameters \cite{brown2020language, liu2019roberta, lepikhin2020gshard, fedus2021switch}. Unfortunately, this need for scaling up comes at the cost of computational parsimony, often a key factor in commercial systems.

One relatively unexplored path toward scaling up language models is via increasing the embedding vocabulary. 
Typically the \textit{embedding} vocabulary (number of rows in the embedding table) is determined by the \textit{token} vocabulary; i.e. there is one embedding for each unique token.
Increasing the embedding vocabulary necessitates increasing the token vocabulary and, undesirably, the softmax parameters. 
Further, often a specific token vocabulary (e.g. a particular 4096-wordpiece vocabulary) is required in order to be compatible with downstream tasks (e.g. for shallow fusion with a speech model which has that vocabulary).

Yet, scaling up the embedding vocabulary has a key advantage: embedding lookups are sparse and lightweight.
For instance in language models, only a single fixed-dimensional vector from the embedding table corresponding to the current token must be retrieved from memory at each step.
In contrast to ``sparse'' parameters such as embeddings, other parameters are ``dense'': they are involved at each step (an exception is conditional computation models, see \S\ref{sec:related}), and consequently must be stored on limited, high-bandwidth GPU/TPU memory.  
Sparse parameters can potentially be stored on off-chip memory, such as on CPU memory, disk, or even a remote parameter server \cite{zhao2020distributed}, and retrieved in a lightweight fashion (typically a few kilobytes per embedding vector).
Since the vocabulary size of the embedding table has no effect on the number of lookups or operations involved in each forward pass, it means that the embedding vocabulary can be scaled up arbitrarily with no increase in compute. 
Practically it is only limited by the device storage capacity, which is typically larger than GPU/TPU memory by a factor of 10 (CPU memory) to 100 (disk), and virtually unlimited for distributed clusters \cite{zhao2020distributed}.

\begin{figure}[t]
  \vspace{-10pt}
  \centering
  \includegraphics[width=\linewidth]{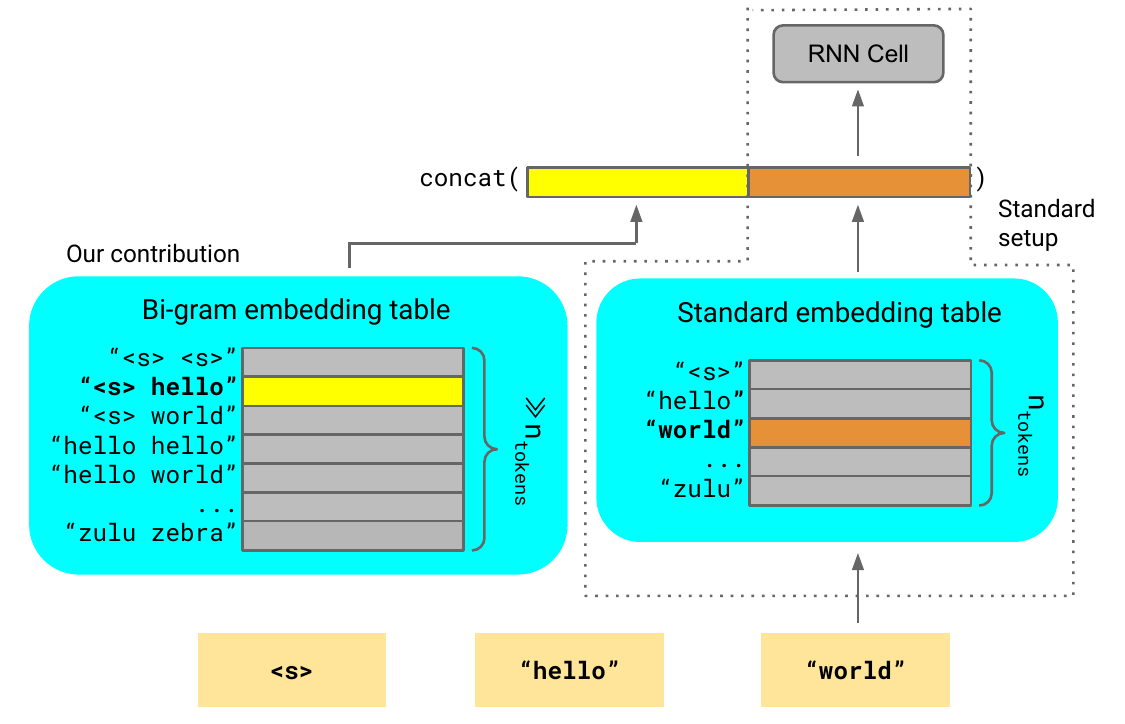}
  \vspace{-15pt}
  \caption{We concatenate an n-gram embedding of the previous tokens (yellow vector) with the input to an RNN cell (red vector). The n-gram embedding table can be scaled to arbitrarily many rows since the number of unique n-grams is exponential.}
  \label{fig:schematic}
  \vspace{-13pt}
\end{figure}

In this work, we present LookupLM, a simple method to scale up the embedding vocabulary in recurrent language models in such a way as to improve modeling quality, \textit{without} changing the token vocabulary.
At each step, LookupLM hashes the sequence of the previous $n$ tokens, draws a vector from a large n-gram embedding lookup table, and concatenates it with the input to the RNN cell, as shown in Figure \ref{fig:schematic}. 
Since the number of possible n-gram sequences grows exponentially with $n$, the n-gram embedding vocabulary size can grow arbitrarily large while commensurately improving its capacity to embed the previous sequence. 
Intuitively this should especially improve modeling of the long tail in subword models, since the n-gram embedding helps with short-range dependencies such as spelling out rare words.

LookupLM achieves a log perplexity decrease of 2.44 over a standard language model with minimal increase in dense model parameters or floating point operations, and it achieves a 23.4\% relative improvement on long tail speech recognition WER when integrated via shallow fusion with an E2E speech model.
We perform various studies to understand the factors that contribute to LookupLM's success.


\subsection{Related Work}
\label{sec:related}

While to our knowledge no work has systematically experimented with scaling up the embedding vocabulary of language models to arbitrarily large sizes (for a fixed token vocabulary), our paper follows several themes of prior publications.
In conditional computation \cite{shazeer2017outrageously}, different blocks of parameters are turned on and off based on each example, but current implementations still place these sparse parameters on GPU/TPU memory \cite{fedus2021switch}.
The idea of integrating an off-chip datastore during inference to improve language modeling was pursued in kNN language models \cite{khandelwal2019generalization}, where a nearest neighbor search is performed in embedding space against a large key-value datastore of (sentence prefix, embedding) pairs rather than an embedding lookup.
Production systems with embedding lookup tables stored on SSD have been demonstrated for ad recommendation with a large number of entities (users, items, etc.) \cite{zhao2020distributed}; we experiment with large embedding vocabularies based on the composition of a \textit{small} number of entities (wordpiece tokens).
There have been a few variations of fast embedding compositions for language decoding. \cite{botha2017natural} and \cite{zhang2019neural} generated word features by summing up a combination of the word's constituent unigrams, bigrams, and trigrams.
An even more general transformation is to use an RNN to consume a token sequence to produce a fixed-length embedding \cite{ling2015character}, but this requires increasing dense parameters. 
Finally the use of n-gram embedding lookups have been demonstrated in RNN-T speech recognition decoders. Both \cite{ghodsi2020rnn} and \cite{variani2020hybrid} showed that feeding into the RNN-T joint network a simple bigram embedding of the previous two tokens, was sufficient to get similar WER as a recurrent prediction network. 
\section{Method}


\subsection{N-gram embedding}
An embedding table is an $U\times E$ matrix where $E$ is the embedding dimension and $U$, the number of embeddings, is typically equal to $V$, the number of unique tokens. 
We wish to disentangle $U$ from $V$ so that that it can be arbitrarily scaled up without affecting the model output space.
Similar to n-gram language models, we make the assumption that the previous $n$-token\footnote{Note that we deviate slightly from standard naming convention; n-gram LMs use the previous $n-1$ tokens rather than $n$.} sequence can be used as a good signal for what the next token should be, and thus desire a way to embed that n-gram.
It would be infeasible to instantiate a unique embedding for every n-gram, since the number of unique n-grams for a token vocabulary $V$ scales exponentially as $V^n$.
Instead, we predetermine an embedding vocabulary size $U$ based on available storage, and assign each n-gram an embedding ID via a modular hash. 
Specifically, the token ID sequence $(t_0, \cdots, t_{n-1})$ is assigned an embedding ID as follows.
\begin{multline}
    \begin{gathered}
        \texttt{ngram2id(}t_0, \cdots, t_{n-1}\texttt{)} = \left(\sum_{i=0}^{n-1}t_iV^i\right) \;\textrm{mod}\; U \\
        \forall \; t_i\in [0, V)
    \end{gathered}
\end{multline}
Modular hashing necessitates collisions; arbitrarily different n-grams will be hashed to the same embedding. As we will show in \S\ref{sec:scale}, the performance improves as we reduce collisions by increasing $U$.

\subsection{Injecting embeddings into the language model}
Once an n-gram embedding vector is obtained, it is concatenated with the input vector before being fed into the RNN cell, as shown in Figure \ref{fig:schematic} for bi-grams.
This causes the RNN to increase dense parameters in order to process the elongated input vector, but the scaling is linear rather than quadratic since the RNN output dimensionality remains fixed. 
We conjecture that the context information from the n-gram embedding would be useful not only at the input to the RNN stack, but at intermediate layers too, just as intermediate RNN cells receive context information via the hidden state specific to that layer. 
Therefore, we inject (concatenate) an n-gram embedding to the input activations of every layer, drawing each from an embedding table specific to that layer.

\subsection{Integrating the LM into the speech model}

For ASR evaluation, the E2E model used is a wordpiece Conformer-encoder \cite{gulati2020conformer}, HAT-factorized \cite{variani2020hybrid} RNN-T decoder \cite{graves2012sequence}.
We integrate our language models with an E2E speech model in two steps: 
(1) obtain an effective E2E likelihood by separating out the speech model's log-posterior from its internal language model score via HAT factorization \cite{variani2020hybrid}:
\vspace{-5pt}

\begin{equation*}
    \log p(x|y) \approx \log p(y|x) - \lambda_2 \log p_{ILM}(y),
\end{equation*}

\vspace{4pt}
\noindent
and (2) add the LM log-posterior score:
\vspace{-12pt}

\begin{equation*}
    y^* = \underset{y}{\mathrm{argmax}}\; \left[\log p(y|x) - \lambda_2 \log p_{ILM}(y) + \lambda_1 p_{LM}(y)\right].
\end{equation*}

\vspace{-5pt}
\noindent
The interpolation weights ($\lambda_1$, $\lambda_2$) are determined via a blackbox optimization service \cite{golovin2017google} where the objective was the WER on a 50:50 mix of voice search queries and TTS-generated sentences with rare proper nouns from the Google Maps domain.


\section{Setup}


\begin{table*}[!h]
    \vspace{-20pt}
    \caption{Main results (a) and ablation studies (b-e)}
    \vspace{-12pt}
    \label{tab:main}
    \centering
    \smallskip\noindent
    \resizebox{.79\linewidth}{!}{%
    \begin{tabular}{ll|ll|lll|lll}
        \toprule
        \multicolumn{1}{c}{} & \multicolumn{1}{c}{} & \multicolumn{2}{c}{\# Params (M)} & \multicolumn{3}{c}{Log perplexity per word} & \multicolumn{3}{c}{WER (\%)} \\
         & Language Model & Dense & Sparse & Head & RareASR & RareALL & Head & RareASR & RareALL \\
        \midrule
        (a) & B0: No LM                   & \gc{0.0}{0.0}{59.5}  & \gc{0.0}{0.0}{805.7}          & -                      & -                      & -                       & \gc{9.6}{6.0}{9.6} & \gc{67.4}{43.5}{67.4} & \gc{77.2}{69.1}{80.3} \\
            & B1: Base LM                 & \gc{5.5}{0.0}{59.5}  & \gc{0.4}{0.0}{805.7}          & \gc{7.09}{6.70}{7.09}  & \gc{17.22}{14.91}{17.22} & \gc{22.62}{20.16}{22.62} & \gc{7.1}{6.0}{9.6} & \gc{56.8}{43.5}{67.4} & \gc{80.3}{69.1}{80.3} \\
            & B2: Large LM                & \gc{59.5}{0.0}{59.5} & \gc{0.4}{0.0}{805.7}          & \gcb{6.70}{6.70}{7.09}  & \gc{14.91}{14.91}{17.22} & \gcb{20.16}{20.16}{22.62} & \gcb{6.2}{6.0}{9.6} & \gc{43.8}{43.5}{67.4} & \gc{71.0}{69.1}{80.3} \\
            & E8: Base LM + Lookup Table  & \gc{9.6}{0.0}{59.5}  & \gc{805.7}{0.0}{805.7}        & \gc{6.71}{6.70}{7.09}  & \gc{14.91}{14.91}{17.22} & \gc{20.18}{20.16}{22.62} & \gc{6.3}{6.0}{9.6} & \gcb{43.5}{43.5}{67.4} & \gcb{69.1}{69.1}{80.3} \\
        \midrule
        (b) & E0: Lookup-4k-2048   & \gc{22.2}{0.02}{22.2} & \gc{25.6}{0.0}{805.7}               & \gc{6.90}{6.70}{7.09} & \gc{15.80}{14.84}{15.95} & \gc{21.40}{20.05}{21.40} & \gc{6.5}{6.0}{6.7} & \gc{50.2}{43.5}{50.2} & \gc{76.9}{69.1}{76.9} \\
            & E1: Lookup-8k-2048   & \gc{22.2}{0.02}{22.2} & \gc{50.7}{0.0}{805.7}               & \gc{6.89}{6.70}{7.09} & \gc{15.69}{14.84}{15.95} & \gc{21.27}{20.05}{21.40} & \gc{6.6}{6.0}{6.7} & \gc{48.3}{43.5}{50.2} & \gc{75.6}{69.1}{76.9} \\
            & E2: Lookup-16k-2048  & \gc{22.2}{0.02}{22.2} & \gc{101.1}{0.0}{805.7}              & \gc{6.83}{6.70}{7.09} & \gc{15.42}{14.84}{15.95} & \gc{20.89}{20.05}{21.40} & \gc{6.5}{6.0}{6.7} & \gc{47.1}{43.5}{50.2} & \gc{73.8}{69.1}{76.9} \\
            & E3: Lookup-33k-2048  & \gc{22.2}{0.02}{22.2} & \gc{201.7}{0.0}{805.7}              & \gc{6.79}{6.70}{7.09} & \gc{15.20}{14.84}{15.95} & \gc{20.56}{20.05}{21.40} & \gc{6.4}{6.0}{6.7} & \gc{45.4}{43.5}{50.2} & \gc{72.6}{69.1}{76.9} \\
            & E4: Lookup-66k-2048  & \gc{22.2}{0.02}{22.2} & \gc{403.0}{0.0}{805.7}              & \gc{6.77}{6.70}{7.09} & \gc{15.05}{14.84}{15.95} & \gc{20.42}{20.05}{21.40} & \gc{6.4}{6.0}{6.7} & \gc{43.7}{43.5}{50.2} & \gc{70.8}{69.1}{76.9} \\
            & E5: Lookup-131k-2048 & \gc{22.2}{0.02}{22.2} & \gc{805.7}{0.0}{805.7}              & \gcb{6.72}{6.70}{7.09} & \gcb{14.89}{14.84}{15.95} & \gcb{20.18}{20.05}{21.40} & \gcb{6.2}{6.0}{6.7} & \gcb{43.5}{43.5}{50.2} & \gcb{69.1}{69.1}{76.9} \\
        \midrule
        (c) & E6: Lookup-131k-2048  & \gc{22.2}{0.0}{22.2} & \gc{805.7}{0.0}{805.7}              & \gc{6.72}{6.70}{7.09} & \gc{14.89}{14.84}{15.95} & \gc{20.18}{20.05}{21.40} & \gc{6.2}{6.0}{6.7} & \gcb{43.5}{43.5}{50.2} & \gcb{69.1}{69.1}{76.9} \\
            & E7: Lookup-262k-1024  & \gc{13.8}{0.0}{22.2} & \gc{805.7}{0.0}{805.7}              & \gcb{6.70}{6.70}{7.09} & \gcb{14.84}{14.84}{15.95} & \gcb{20.05}{20.05}{21.40} & \gcb{6.0}{6.0}{6.7} & \gc{44.1}{43.5}{50.2} & \gc{70.3}{69.1}{76.9} \\
            & E8: Lookup-524k-512   & \gc{9.6}{0.0}{22.2}  & \gc{805.7}{0.0}{805.7}              & \gc{6.71}{6.70}{7.09} & \gc{14.91}{14.84}{15.95} & \gc{20.18}{20.05}{21.40} & \gc{6.3}{6.0}{6.7} & \gcb{43.5}{43.5}{50.2} & \gcb{69.1}{69.1}{76.9} \\
            & E9: Lookup-1049k-256  & \gc{7.5}{0.0}{22.2}  & \gc{805.7}{0.0}{805.7}              & \gc{6.71}{6.70}{7.09} & \gc{14.99}{14.84}{15.95} & \gc{20.30}{20.05}{21.40} & \gc{6.2}{6.0}{6.7} & \gc{44.4}{43.5}{50.2} & \gc{69.9}{69.1}{76.9} \\
            & E10: Lookup-2097k-128 & \gc{6.5}{0.0}{22.2}  & \gc{805.7}{0.0}{805.7}              & \gc{6.75}{6.70}{7.09} & \gc{15.12}{14.84}{15.95} & \gc{20.50}{20.05}{21.40} & \gc{6.3}{6.0}{6.7} & \gc{44.9}{43.5}{50.2} & \gc{69.8}{69.1}{76.9} \\
            & E11: Lookup-4194k-64  & \gc{6.0}{0.0}{22.2}  & \gc{805.7}{0.0}{805.7}              & \gc{6.88}{6.70}{7.09} & \gc{15.57}{14.84}{15.95} & \gc{21.16}{20.05}{21.40} & \gc{6.5}{6.0}{6.7} & \gc{46.5}{43.5}{50.2} & \gc{72.4}{69.1}{76.9} \\
            & E12: Lookup-8389k-32  & \gc{5.7}{0.0}{22.2}  & \gc{805.7}{0.0}{805.7}              & \gc{6.83}{6.70}{7.09} & \gc{15.61}{14.84}{15.95} & \gc{21.00}{20.05}{21.40} & \gc{6.4}{6.0}{6.7} & \gc{48.2}{43.5}{50.2} & \gc{73.1}{69.1}{76.9} \\
            & E13: Lookup-16777k-16 & \gc{5.6}{0.0}{22.2}  & \gc{805.7}{0.0}{805.7}              & \gc{6.89}{6.70}{7.09} & \gc{15.95}{14.84}{15.95} & \gc{21.32}{20.05}{21.40} & \gc{6.7}{6.0}{6.7} & \gc{48.9}{43.5}{50.2} & \gc{74.8}{69.1}{76.9} \\
        \midrule
        (d) & E14: Lookup-33k-2048-2 & \gc{22.2}{0.02}{22.2} & \gc{201.7}{0.0}{805.7} & \gcb{6.78}{6.70}{7.09} & \gc{15.23}{14.84}{15.95} & \gc{20.65}{20.05}{21.40} & \gcb{6.2}{6.0}{6.7} & \gc{46.6}{45.4}{50.2} & \gc{74.1}{69.1}{76.9} \\
            & E15: Lookup-33k-2048-3 & \gc{22.2}{0.02}{22.2} & \gc{201.7}{0.0}{805.7} & \gc{6.79}{6.70}{7.09} & \gc{15.24}{14.84}{15.95} & \gc{20.66}{20.05}{21.40} & \gcb{6.2}{6.0}{6.7} & \gc{46.4}{45.4}{50.2} & \gc{73.4}{69.1}{76.9} \\
            & E16: Lookup-33k-2048-4 & \gc{22.2}{0.02}{22.2} & \gc{201.7}{0.0}{805.7} & \gc{6.79}{6.70}{7.09} & \gcb{15.20}{14.84}{15.95} & \gcb{20.56}{20.05}{21.40} & \gc{6.4}{6.0}{6.7} & \gcb{45.4}{45.4}{50.2} & \gcb{72.6}{69.1}{76.9} \\
            & E17: Lookup-33k-2048-5 & \gc{22.2}{0.02}{22.2} & \gc{201.7}{0.0}{805.7} & \gc{6.79}{6.70}{7.09} & \gc{15.24}{14.84}{15.95} & \gc{20.67}{20.05}{21.40} & \gc{6.3}{6.0}{6.7} & \gc{46.7}{45.4}{50.2} & \gc{73.8}{69.1}{76.9} \\
            & E18: Lookup-33k-2048-6 & \gc{22.2}{0.02}{22.2} & \gc{201.7}{0.0}{805.7} & \gc{6.80}{6.70}{7.09} & \gc{15.22}{14.84}{15.95} & \gc{20.66}{20.05}{21.40} & \gc{6.4}{6.0}{6.7} & \gc{45.5}{45.4}{50.2} & \gc{72.8}{69.1}{76.9} \\
        \midrule
        (e) & E19: Lookup-33k-2048         & \gc{22.2}{0.0}{22.2} & \gc{201.7}{0.0}{805.7} & \gc{6.79}{6.70}{7.09} & \gc{15.20}{14.84}{15.95} & \gc{20.56}{20.05}{21.40} & \gc{6.4}{6.0}{6.7} & \gc{45.4}{45.4}{50.2} & \gc{72.6}{69.1}{76.9} \\
            & E20: + Incl. curr. token     & \gc{22.2}{0.0}{22.2} & \gc{201.7}{0.0}{805.7} & \gc{6.78}{6.70}{7.09} & \gc{15.16}{14.84}{15.95} & \gc{20.58}{20.05}{21.40} & \gc{6.3}{6.0}{6.7} & \gc{45.6}{45.4}{50.2} & \gc{72.6}{69.1}{76.9} \\
            & E21: + Layer0 injection only & \gc{9.6}{0.0}{22.2}  & \gc{67.5}{0.0}{805.7}  & \gc{6.84}{6.70}{7.09} & \gc{15.40}{14.84}{15.95} & \gc{20.95}{20.05}{21.40} & \gc{6.5}{6.0}{6.7} & \gc{46.3}{45.4}{50.2} & \gc{74.3}{69.1}{76.9} \\
        \bottomrule
    \end{tabular}}
    \vspace{-8pt}
\end{table*}

\subsection{Model}
\label{sec:model}
Our baseline is a 4096-wordpiece RNN language model implemented in Lingvo on Tensorflow
\cite{shen2019lingvo, lingvolm}.
Its architecture is a 2-layer, 512-width, layer-normalized LSTM with an input embedding (not to be confused with the n-gram embedding) dimensionality of 96. When not explicitly stated otherwise, our n-gram embeddings have a dimensionality of 2048, and $n$ is 4.

\subsection{Training}
\label{sec:training}
Our LM training dataset, identical to that of \cite{peyser2020improving}, is a sample of anonymized Google Maps queries, preprocessed with misspelling removal against a whitelist of 1 million common words, and with $\log(n)$ scaling of sentence frequencies to ensure representation on the tail. 
We trained for 300k steps using the Adam optimizer with exponentially decaying learning rate schedule and a batch size of 16384  across 64 TPUv3 accelerator chips.
For ASR evaluation, our RNN-T model was trained on 290M audio-text pairs spanning the domains of search, farfield, telephony, and YouTube \cite{narayanan2019recognizing}.
All parameters, including the embedding tables, are initialized from scratch; we do not load pretrained embeddings.

\subsection{Evaluation}
\label{sec:evaluation}
We report language model log perplexity per word as well as speech recognition word error rate (WER) on 3 test sets of 10,000 sentences from the same domain as the training data but curated in different ways:
\begin{itemize}[leftmargin=*]
    \item Head: simply a held-out part of the LM training set, measuring overall accuracy. 
    \item RareASR: held-out sentences containing a word that is ``rare''---which we henceforth define as appearing \textit{at most} 5 times---in the ASR data but not rare in the LM data. It measures the ability of LM integration to help ASR models learn words that were absent or infrequent in the speech data. 
    \item RareALL: held-out sentences containing a word that is rare in both the ASR and LM data. It measures accuracy on the long tail.
\end{itemize}
For each sentence in RareASR and RareALL, we report LogPP and WER on only the rare word portion of the sentence---even though the entire sentence is passed to the model---in order to extract tail performance.
For ASR evaluations, we synthesize TTS audio for each of the test sets. We have separately found that performance on TTS audio is correlated with recorded audio and thus believe that TTS audio is sufficient for measuring the relative performance gains of the LookupLM architecture.

\section{Results}

\begin{table}[b]
    \vspace{-12pt}
    \caption{Decoding samples picked from the top 5 wins or losses.}
    \vspace{-12pt}
    \label{tab:decodes}
    \centering
    \smallskip\noindent
    \resizebox{.99\linewidth}{!}{%
    \begin{tabular}{ll|ll}
        \toprule
        \multicolumn{2}{c}{LookupLM losses} & \multicolumn{2}{c}{LookupLM wins} \\
        Base LM & LookupLM & Base LM & LookupLM \\
        \midrule
        \begin{tabular}[c]{@{}l@{}}\textcolor{ForestGreen}{Funway}\\\textcolor{ForestGreen}{Foxboro}\end{tabular} & \begin{tabular}[c]{@{}l@{}}\textcolor{red}{fun way}\\\textcolor{red}{foxborough}\end{tabular} & \begin{tabular}[c]{@{}l@{}}\textcolor{red}{they would weigh}\\Louisville Kentucky\end{tabular} & \begin{tabular}[c]{@{}l@{}}\textcolor{ForestGreen}{faywood Way}\\Louisville Kentucky\end{tabular} \\
        \midrule
        \begin{tabular}[c]{@{}l@{}}yearly \textcolor{ForestGreen}{Ball}\\\textcolor{ForestGreen}{Des Moines}\end{tabular} & \begin{tabular}[c]{@{}l@{}}yearly\\\textcolor{red}{Baltimore}\end{tabular} & \textcolor{red}{I'll go away} & \textcolor{ForestGreen}{Algoa Bay} \\
        \bottomrule
    \end{tabular}}
    \vspace{-15pt}
\end{table}


\subsection{Comparison to baselines}
\label{sec:baseline}

\begin{table*}[t]
    \vspace{-15pt}
    \caption{Multidomain results.}
    \vspace{-12pt}
    \label{tab:multidomain}
    \centering
    \smallskip\noindent
    \resizebox{.83\linewidth}{!}{%
    \begin{tabular}{l|ll|llllll}
        \toprule
        \multicolumn{1}{c}{} & \multicolumn{2}{c}{\# Params (M)} & \multicolumn{6}{c}{WER (\%)} \\
        Language Model & \# Dense & \# Sparse & Voice Search & Maps & News & Play & Search & YouTube \\
        \midrule
        B3: Multidomain LM                 & 126.7 & 0.4  & 6.1 & 19.8 & 9.0 & 45.3 & 32.5 & 28.0 \\
        E22: Multidomain LM + Lookup Table & 135.1 & 67.5 & 6.1 & \textbf{19.4} & \textbf{8.8} & \textbf{44.6} & \textbf{32.0} & \textbf{27.6} \\
        \bottomrule
    \end{tabular}}
    \vspace{-10pt}
\end{table*}

In Table \ref{tab:main}a, we evaluate language model log perplexity and downstream WER against the number of dense and sparse parameters it requires.
Our best LookupLM model (E8: Base LM + Lookup Table) contains a embedding vocabulary of 524k and dimensionality of 512. We compare its results to 3 baselines:
\begin{itemize}[leftmargin=*]
    \item B0: No LM - ASR predictions without LM integration.
    \item B1: Base LM - The default LM described in \S\ref{sec:model} without n-gram embeddings. It has the same architecture as LookupLM minus the lookup table and is the main baseline.
    \item B2: Large LM - Baseline B1 but with 10x more parameters by having 2048 input/output activations instead of 512. It shows how much one would need to increase dense parameters in order to achieve similar results to LookupLM.
\end{itemize}
All models significantly outperform B0, indicating the ASR model is ripe for LM integration. 
LookupLM (E8) achieves noticeable gains compared to Base LM (B1).
On the Head test set, the log perplexity decreases by 0.8 and the WER by 11.2\% relative to the base. 
On the RareALL test set, log perplexity decreases by 2.44 and WER by 13.9\% relative.
Finally, on the RareASR test set the WER decreases by 23.4\% relative, showing that LookupLM is an effective architecture for improving long tail speech recognition with sufficient text-only data. Table \ref{tab:decodes} shows some decoding samples.

We now compare with Large LM (B2) to see how much conventional model scaling is required to achieve the same gains as LookupLM. 
B2 has similar performance as LookupLM but requires 6.2 times more dense parameters and hence 6.2$\times$ the FLOPs, showing that using a large, off-chip lookup table can effectively trim on-chip model size and reduce computation.

\subsection{Scaling of embedding vocabulary}
\label{sec:scale}
We now study model performance scaling with embedding vocabulary size through an ablation study in Table \ref{tab:main}b, which scales the vocabulary from a size of 4k (E0) to 131k (E5).
While the embedding vocabulary can be much larger if stored off-TPU, for this demonstration we stored the embedding table on TPU and thus 131k was the limit of available memory at 2048-dimensional embeddings.
We name each model according to the template, Lookup-\$VOCAB-\$DIM, where \$VOCAB is the embedding vocabulary size and \$DIM is the embedding dimensionality.
The performance increases with embedding vocabulary, with no sign of saturating. The scaling of relative gain in this range roughly follows a power law with an exponent of $\sim$0.1 for linear perplexity and $\sim$0.5 for WER.

\subsection{Vocabulary size vs. embedding dimension tradeoff}
Given a fixed amount of sparse parameters, there is a tradeoff between the embedding vocabulary and embedding dimension. 
Table \ref{tab:main}c shows various LookupLM models with a fixed number of sparse parameters but differing amounts of rows and columns. 
The results show a sweet spot around E6-E8, which have embedding dimensionalities above 256. Among these, we pick E8 as our best model since it requires the fewest dense parameters.

\subsection{Effect of n-gram order}
The n-gram order, or simply $n$, determines the context size for the embedding. A larger n-gram order entails a larger context, but also results in more collisions given that the number of possible n-grams increases with $n$. Increasing n-gram order also reduces the number of times any particular n-gram is updated.
We aim to look for a sweet spot in Table \ref{tab:main}d, where the naming template is Lookup-\$VOCAB-\$DIM-\$ORDER. We find that there is only a slight variation in performance with n-gram order for values between 2 (E14) and 6 (E18), with 4 (E16) slightly being the best.

\subsection{Design space variations}
Finally, we consider two variations of our model architecture:
\begin{itemize}[leftmargin=*]
    \item E20: Instead of the n-gram embedding entailing the previous $n$ tokens, we have it entail the current token plus the previous $n-1$ tokens.  
    \item E21: Instead of injecting the n-gram embedding at every layer, we inject it only at the first layer.
\end{itemize}

Table \ref{tab:main}e shows that shifting the context window of the n-gram embedding to include the current token (E20) makes no difference compared to not shifting (E19). It's advantageous to not include the current token because then the n-gram embedding can be pre-fetched before the current step, reducing latency. Removing the multi-layer injection (E21) reduces performance slightly, but also reduces the number of dense (and sparse) parameters significantly. Exploring the tradeoffs between single- vs. multi-layer injection may lead to a more optimal setup in terms of performance and parameter count, but we leave that to future work.

\subsection{LookupLM maximizes resource utilization}
We have separated \textit{dense} and \textit{sparse} parameters throughout this paper because they represent the more fundamental time and space complexities. 
In an LSTM, the dense parameters are each accessed once per step, so the number of dense parameters is directly correlated with the number of floating point operations (FLOPs), i.e. time complexity. 
Meanwhile, the storage requirement, i.e. space complexity, is dominated by the sparse parameters (increasing dense parameters also increases memory, but by a relatively much smaller amount). 
In typical networks, the time and space complexity are inextricably linked: improving performance requires increasing both parameter count \textit{and} FLOPs. 
LookupLM provides a way of achieving performance by increasing memory without increasing FLOPs. 
This can allow for taking advantage of excess resources (typically RAM or disk memory).
Table \ref{tab:tradeoff} shows the WER for a grid of independent memory and FLOPs configurations for LookupLM, which can be useful for determining the tradeoff between various resource allocations.
Performance relative to Base LM (B1) improves by shown amounts when resources along either axis (memory or FLOPs) are increased.

\begin{table}[!h]
    \vspace{-2pt}
    \caption{Relative WER (\%) on RareASR at various complexities.}
    \vspace{-12pt}
    \label{tab:tradeoff}
    \centering
    \smallskip\noindent
    \resizebox{.99\linewidth}{!}{%
    \begin{tabular}{ll|lllll}
        \toprule
                  & \small \# FLOPs (M) &             19.3  &             38.8 &             65.5 &             99.2 &             140.0 \\
        Mem.      & \small \# Dense (M) &             9.6  &             19.4 &             32.8 &             49.6 &             70.0 \\
        (GB) & \small \# Sparse (M) &                  &                  &                  &                  &                  \\
        \midrule
        0.2         & 50.7           &  \gradient{-15.9} &  \gradient{-20.7} &  \gradient{-21.2} &  \gradient{-24.5} &  \gradient{-23.4} \\
        0.4         & 101.1          &  \gradient{-18.0} &  \gradient{-20.3} &  \gradient{-21.6} &  \gradient{-25.1} &  \gradient{-24.9} \\
        0.8         & 201.7          &  \gradient{-19.7} &  \gradient{-21.3} &  \gradient{-24.2} &  \gradient{-24.3} &  \gradient{-24.2} \\
        1.6         & 403.0          &  \gradient{-22.5} &  \gradient{-23.9} &  \gradient{-24.8} &  \gradient{-23.9} &  \gradient{-26.4} \\
        3.2         & 805.7          &  \gradient{-24.0} &  \gradient{-23.3} &  \gradient{-25.2} &  \gradient{-24.3} &  \gradient{-26.5} \\
        \bottomrule
    \end{tabular}}
    \vspace{-7pt}
\end{table}

\subsection{Preliminary multidomain results}
Table \ref{tab:multidomain} shows preliminary results on a production-scale LSTM LM (4-layer, 2048-width) trained for 1M steps on $\sim$200B text examples from the domains of Voice Search, Maps, News, Play, Search, and YouTube. 
Besides Voice Search, all evaluation test sets in Table \ref{tab:multidomain} are curated to measure long tail performance.
With a modest Lookup Table size (67.5M parameters), we achieve modest WER gains across all the domains on the long tail, without harming Voice Search performance.

\section{Conclusion}
This paper was a first exploration into scaling up language models via large embedding tables. 
While our method here is already effective, it can still be improved. For example, since it's clear that decreasing collisions helps performance, we could assign the most common n-grams a unique embedding ID while modular hashing the rest. 
Further,
it may help to apply a per-embedding learning rate schedule which is based on the number of times each embedding has been updated rather than on the global step.
We leave these ideas as future work.

\section{Acknowledgements}
We thank Bo Li, Ruoming Pang, and Chen Zhu for helpful discussions and technical guidance.

\newpage

\bibliographystyle{IEEEtran}
\bibliography{bibliography}

\end{document}